\documentclass[letterpaper, 10 pt, conference]{ieeeconf}
\IEEEoverridecommandlockouts
\usepackage{cite}
\usepackage{amsmath,amssymb,amsfonts}
\usepackage{algorithmic}
\usepackage{graphicx}
\usepackage{textcomp}
\usepackage{xcolor}
\usepackage{balance}
\usepackage{epstopdf}
\usepackage[normalem]{ulem} 
\usepackage{upgreek}
\usepackage{comment}
\usepackage{url}

\def\BibTeX{{\rm B\kern-.05em{\sc i\kern-.025em b}\kern-.08em
    T\kern-.1667em\lower.7ex\hbox{E}\kern-.125emX}}
\let\oldhat\hat
\renewcommand{\vec}[1]{\boldsymbol{#1}}
\renewcommand{\hat}[1]{\oldhat{\mathbf{#1}}}
\newcommand{\mat}{\vec}

\title{\LARGE \bf Error Decomposition for Hybrid Localization Systems}

\author{
    Benedict Flade$^{1*}$, Simon Kohaut$^{2*}$, Julian Eggert$^{1}$
    \thanks{* Authors contributed equally}%
    \thanks{$^{1}$ Honda Research Institute Europe GmbH,
    	    Carl-Legien-Str. 30,\newline\hspace*{1.6em} 63073 Offenbach, Germany \newline\hspace*{1.6em}
            {\tt\small firstname.surname@honda-ri.de}}%
    \thanks{$^{2}$ Technical University of Darmstadt, 64283 Darmstadt, Germany}%
}

\usepackage{tikz}
\usepackage[color=black,opacity=1,angle=0,scale=1]{background}
\backgroundsetup{
  contents={
      \begin{tikzpicture}
        \node at (current page.center) [align=center] {\textcopyright 2021 IEEE. Personal use of this material is permitted. \\ Permission from IEEE must be obtained for all other uses, in any current or future media, including reprinting/republishing this material for advertising or promotional purposes, \\ creating new collective works, for resale or redistribution to servers or lists, or reuse of any copyrighted component of this work in other works.
         \\ DOI:10.1109/ITSC48978.2021.9564415       };
    \end{tikzpicture}},
  placement=bottom,
  scale=0.6,
  vshift=20
}

\begin{document}

\maketitle
\thispagestyle{empty}
\pagestyle{empty}

\begin{abstract}
    Future advanced driver assistance systems and autonomous vehicles rely on accurate localization, which can be divided into three classes: 
a) viewpoint localization about local references (e.g., via vision-based localization), b) absolute localization about a global reference system (e.g., via satellite navigation), and c) hybrid localization, which presents a combination of the former two. 
Hybrid localization shares characteristics and strengths of both absolute and viewpoint localization. 
However, new sources of error, such as inaccurate sensor-setup calibration, complement the potential errors of the respective sub-systems. 
Therefore, this paper introduces a general approach to analyzing error sources in hybrid localization systems.
More specifically, we propose the Kappa-Phi method, which allows for the decomposition of localization errors into individual components, i.e., into a sum of parameterized functions of the measured state (e.g., agent kinematics).
The error components can then be leveraged to, e.g., improve localization predictions, correct map data, or calibrate sensor setups.  
Theoretical derivations and evaluations show that the algorithm presents a promising approach to improve hybrid localization and counter the weaknesses of the system's individual components.

\end{abstract}

\section{Introduction}

Accurate localization is one of the core challenges of advanced driver assistance systems (ADASs) and autonomous vehicles (AVs).  
In general, localization can be understood as the task of determining a state  (position, orientation, etc.) relative to local and global entities of reference.
Local entities of reference are objects and structures of the surrounding environment.
This includes tangible objects such as road curbs, traffic signs, vehicles, lane markings, and buildings, as well as intangible structures, e.g., the center of a road.
On the other hand, global entities of reference are physical or intangible objects that place no requirement of proximity while being universally available, with the geographical center of Earth being one example.
With this in mind, we separate the localization task into three classes: a)~viewpoint localization, b)~absolute localization, and c)~hybrid localization.

\section{Related work}
\label{sec:related_work}

\begin{figure}
    \centering
    \includegraphics[width=1\linewidth]{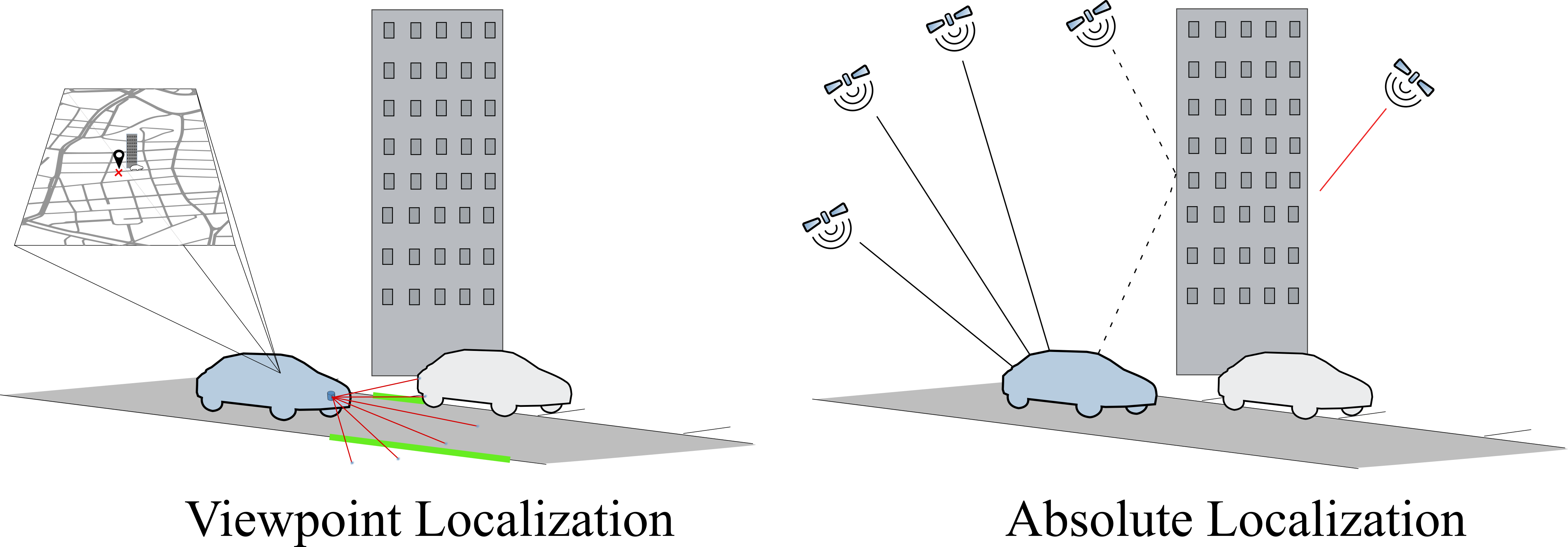}
    \caption{Comparison of viewpoint and absolute localization. Absolute localization describes the ego-position with regard to a global reference, while viewpoint localization relies on matching local observations of the environment with a knowledge base (e.g., map).}
    \vspace{-0.5cm}
    \label{fig:absolute_viewpoint}
\end{figure}

With the term \textbf{viewpoint localization}, we summarize methods that rely on observations of the local environment (see Figure~\ref{fig:absolute_viewpoint}, left). 
Many viewpoint localization approaches employ exteroceptive sensors such as cameras, LiDAR, or radar to observe the surroundings and localize within.
State-of-the-art methods use such sensors to match extracted features to an internal model of the environment \cite{OPink2008, ASchindler2013}.
Generally, the environment can be described as a local map containing topological, semantic, and geometric information.
One example of solving the localization task is to compare features of a camera image with road geometry features of the map, referred to as camera-to-map alignment \cite{Flade2018}.
Potential maps range from 3D point clouds~\cite{Caselitz2016} to polyline-level road shapes~\cite{Hu2004}, with road markings presenting one of the most commonly used entities of reference~\cite{Lu2017,Flade2020}.
An advantage of viewpoint localization is that it presents the natural possibility to validate the plausibility of the observations and relations based on the underlying environment model.
Due to the perception-based nature, viewpoint localization furthermore lends itself to determining the orientation.
However, common challenges of viewpoint approaches are occlusions of references, moving or disappearing references, weather changes, and seasonal effects.
Furthermore, the accuracy and universality of position estimates depend on the quality of the sensors and environment model utilized, i.e., the map. 
Incorrect sensor calibration or map errors (e.g., offsets) may generate consistent and locally accurate results within a particular frame.
Still, they cannot be exchanged with other agents that do not share the same sensors or environment model.

On the other hand, there is \textbf{absolute localization}, which describes an entity with regard to a global reference frame (see Figure~\ref{fig:absolute_viewpoint}, right).
Especially in situations in which agents traverse the world freely and are not bound to a specific local frame, such universal reference is valuable.
A common example of this is Global Navigation Satellite Systems (GNSS), which includes GPS, GLONASS, Galileo, and BeiDou. 
GNSSs realize such universal reference by describing positions with regard to an Earth-centered Earth-fixed Cartesian coordinate system or ellipsoid-based spherical coordinate system. 
To improve the GNSS signal, proprioceptive vehicle sensors observing internal states (e.g., acceleration) are used to form an inertial navigation system (INS)~\cite{Rezaei2007}. 
Compared to viewpoint localization, which exploits a self-generated or acquired environment representation, absolute localization directly determines the position, e.g., by trilateration, without the restraint of requiring knowledge of the environment.
However, using no local information also means no conclusions, e.g., about the current lane, can be drawn.
Furthermore, effects caused by the environment, e.g., multipath errors in urban canyons, degrade the localization performance nonetheless \cite{Zhu2018}. 

As a third class, \textbf{hybrid localization} combines absolute and viewpoint localization, see, e.g.,~\cite{Krakiwsky1988} or~\cite{Toledo2009}.
In general, hybrid localization benefits from the strengths of the individual characteristics of its components. 
Yet, potential sources of error, such as unwanted offsets or distortions of the reference frames, also accumulate.
Lu et al.\ show a hybrid system that fuses GNSS signals, vision sensors, and open-source map data~\cite{Lu2014}.
The authors recognize that such systems strongly depend on correct map data, successful vision-based feature detection, and correct interpretation of the data.
For this reason, Lu et al.\ introduce a probabilistic error model that considers possible errors.
Individual error models for lane detection, localization, and map topology are introduced in this context.
Hereby, the paper focuses on failure detection, i.e., distinguishing faulty and correct behavior based on defined thresholds.
Examples of further research that addresses error analysis are~\cite{Brunker2017} for handling GNSS shortages,~\cite{Dong2016} for tackling GNSS-related multipath errors, or~\cite{Murphy2019} for detecting map errors.
However, in all those cases, errors are considered individually.

Research from the field of absolute localization that attempts to correct errors of different sources has been presented in the context of map matching~\cite{Liu2008} and dead reckoning~\cite{Betaille2000}.
Similarities between map-based reference paths and GPS are leveraged to match one to the other and improve a vehicle's position estimates.

We believe that considerable potential for error correction lies within the combination of absolute and viewpoint localization.
The analysis of a disparity that arises from comparing the two separate localization systems can be exploited to identify quantitative, semantic, and geometric errors.
Therefore, we propose to leverage the characteristics of a hybrid system to simultaneously estimate and decompose errors of multiple sources.
This paper presents the novel Kappa-Phi ($\vec{\kappa}\vec{\phi}$) method that leverages the outputs of a dynamic hybrid localization system.
The goal is to estimate systematic localization errors that are only identifiable due to the combination of disparate localization processes.
This is done by anticipating and correcting disparities between the position estimates of the subcomponents, which operate within different frames and are based on disjoint observation spaces. 
More specifically, we introduce a general model to simultaneously decompose localization errors into a sum of functions of the agent's kinematic states, as reported by the individual localization systems. 
While the Kappa-Phi method for error decomposition is not limited to a specific learning technique, we choose the Unscented Kalman Filter for our proposed implementation.
The concept chooses the error space as measurement space and estimate of error parameters instead of the measurement of the actual position.

What separates our research from preliminary work is the online investigation of localization error sources without a strict requirement of fixed assumptions on the system setup.
This means that the concept allows system characteristics, e.g., sensor positions, to be initially unknown or incorrect as long as the independence of the localization components is guaranteed.

The remainder of this paper is structured as follows. 
Section~\ref{sec:methods} presents fundamental design principles and goes into more detail on hybrid localization systems. 
This is followed by introducing the novel error decomposition concept, including its corresponding error models in section~\ref{sec:error_models}.
Potential applications for leveraging the gained knowledge and experiments that show simulation results are outlined in section~\ref{sec:error_application}.  
The paper is rounded off in section~\ref{sec:conclusion} by a conclusion and an outlook.

\section{Fundamental Considerations}
\label{sec:methods}

\begin{figure*}[ht!]
    \centering
	\includegraphics[width=0.8\linewidth]{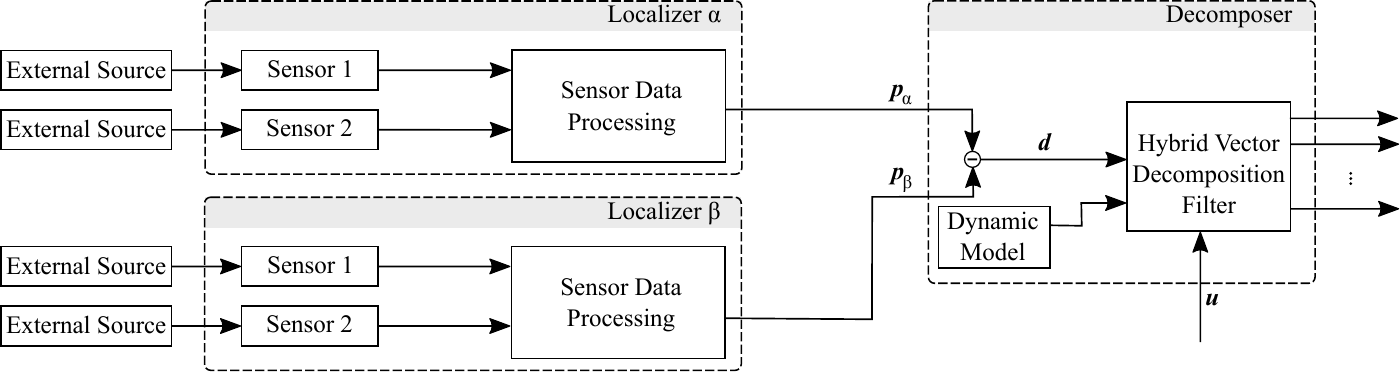}
    \caption{Hybrid Localization System consisting of two localizers and an error decomposition component. Potential external sources of the localizers include ephemeris correction data for GNSS receivers or map geometry data for viewpoint localization. The inputs of the decomposition filter are the estimates of the localizer as well as an input vector $\vec{u}$ that describes measured states (e.g., heading angle or time), according to the chosen error model.}
    \label{fig:hybrid_loc}
    \vspace{-0.4cm}
\end{figure*}

\subsection{Design Decisions}

Localization is a crucial component of a wide range of technologies and applications. 
However, all methods are prone to individual errors.
We therefore target a \textbf{general} approach that allows for a plausibilization of errors and prediction of the error evolution.
With regard to the nature of the error, the concept shall be capable of handling different error types depending on the application, while being \textbf{easily extendable}. 
The plausibilizations shall be on a quantitative level, i.e., leading to \textbf{explicit values} in the unit of the measurement.

Concerning \textbf{real-world application}, we target a solution that is suitable for usage in state-of-the-art production vehicle setups without the restraint of being limited to a predefined sensor setup. 
While accepting a potential initialization or learning phase, we expect the process to be handled during usual behavior without requiring specific maneuvers, movements, or boundary conditions.
Furthermore, a plausibilization of multiple components shall be possible simultaneously, given certain observability requirements are met.

\subsection{Hybrid Localization Systems}
\label{sec:hybrid_loc_sys}

In general, the term hybrid indicates that a system consists of two or more subsystems.
In the broad sense of localization, this generally applies to systems of heterogeneous components with, e.g., an inertial measurement unit (IMU) and GNSS receiver.
However, in view of the task of error decomposition, we narrow down the term hybrid localization to describe an approach based on two independent localization concepts, e.g., satellite-based and vision-based.

Fig. \ref{fig:hybrid_loc} shows a hybrid localization system as described above, consisting of two localizers $\upalpha$ and $\upbeta$.
We stress that each localization component can consist of a single or multiple individual and heterogeneous sensors.
In the latter case, we assume a logical center of the sensor set to be defined, to anchor the setup in the vehicle.

To efficiently leverage the localization components, hereinafter referred to as localizers, we need to define further requirements.
First, the \textbf{localizers need to be independent} without communication between each other so that intrinsic errors of the sensors are not adopted or shared.
This also means that the localizers are sensitive to different error domains.
Second, we expect each localizer to provide a \textbf{separate location estimate} with regard to its corresponding reference frame.
Additionally, an initial assumption of the relation between the reference frames, i.e., the transformation from one reference frame into another, is beneficial.
One common example is the alignment of two reference frames with different origins. 

\subsection{Error Decomposition}

The Kappa-Phi method decomposes errors of a hybrid localization system into a sum of parameterized functions of the agent's kinematic state.
More specifically, it observes and identifies dynamic system states and parameters of systematic errors.
The resulting insights can be used to improve localization predictions on the one hand, as well as correcting the sources of the error.

\subsubsection{Difference Vector}

As outlined in \ref{sec:hybrid_loc_sys}, we consider a hybrid localization system with two individual localizers that each provide state estimates with regard to the respective reference frame.
Note that the two localizers may consist of several rigidly connected heterogeneous sensors with one pre-defined logical center.
For simplicity, we suggest placing the logical center of both localizers at the same location, e.g., in one of the sensors or in the center of the rear axis of a vehicle.
At their best, both localizers $\upalpha$ and $\upbeta$ provide a correct estimate.
However, in real-world applications, several aspects impact successful localization, resulting in ambiguous localization estimations.
Examples include errors in the individual sensors, errors in the assumption of the environment, and calibration errors of the sensor setup.

Based on an initial guess on the relation of the reference frames, the first step is to align the reference frames of both estimates  $\vec{p}_\upalpha$ and $\vec{p}_\upbeta$, e.g., by translation.
The superposition of all occurring errors causes a disparity between the localization estimates $\vec{p}_\upalpha$ and $\vec{p}_\upbeta$.
This disparity is described by the difference vector $\vec{d}$ according to 
\begin{equation}
    \label{eq:diff_vector_measured}
	\vec{d} = \vec{p}_\upalpha - \vec{p}_\upbeta.
\end{equation}
In general, the difference vector can be described as an offset value, e.g., planar or spatial, that qualifies for a mismatch between heterogeneous sources of information.
We propose identifying systematic errors by decomposing the vector into a sum of parameterized functions of the agent's kinematic state. 

\subsubsection{Difference Model}

The disparity between the localization estimates $\vec{p}_\upalpha$ and $\vec{p}_\upbeta$ is caused by one or multiple error sources and is influenced by the kinematic state $\vec{u}$ of the agent.
A difference vector that changes its length and orientation when the agent moves through its environment makes the latter visible.

While the difference vector $\vec{d}$ is a real-world observation, we now formulate a model that mathematically describes this difference, hereinafter referred to as difference model~$\operatorname{D}$.
More specifically, we define a function $\operatorname{D}(\vec{x}, \vec{u})$ that models the localization difference by considering individual error sources as a sum of  error components $\vec{\phi}^i$ which are dynamic vectors depending on the vehicle state, plus an error component $\vec{\kappa}$ that is independent of the agent's  state $\vec{u}$.
For the difference model, we formulate
\begin{equation}
	\operatorname{D}(\vec{x}, \vec{u}) = \vec{\kappa}(\vec{x}) + \sum_i \vec{\phi}^i(\vec{x}, \vec{u}).
	\label{eq:diff_vec}
\end{equation}

\begin{figure}[hb!]
    \centering
	\includegraphics[width=0.65\linewidth]{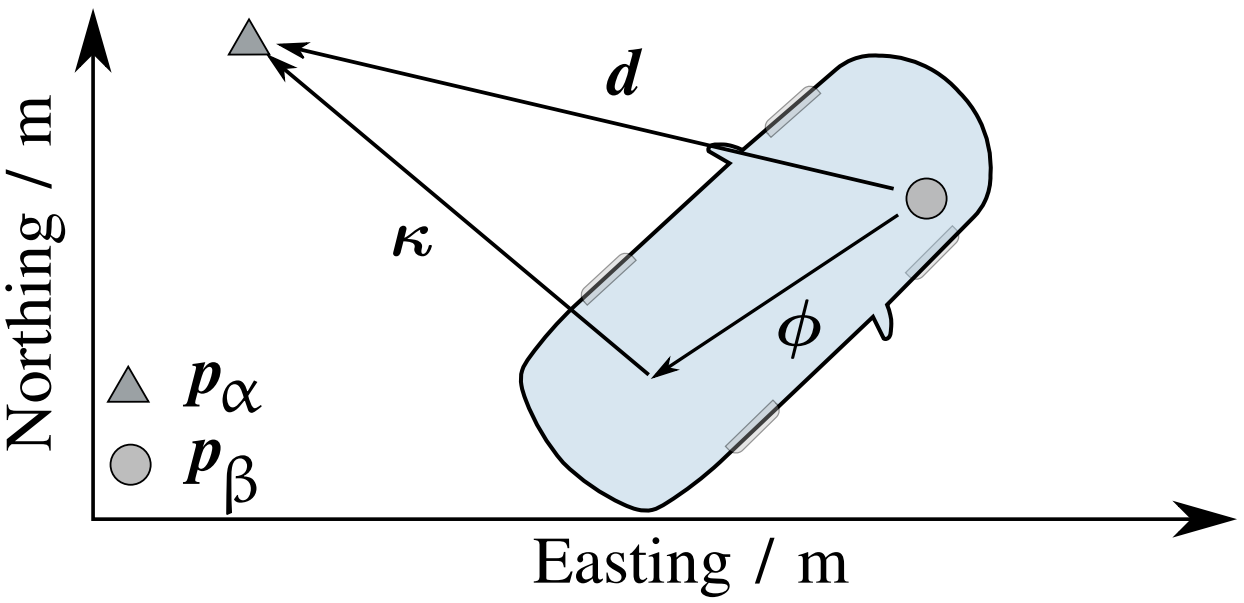}
    \caption{Example of two erroneous position estimates $\vec{p}_\upalpha$ and $\vec{p}_\upbeta$ (obtained from two localizers $\upalpha$ and $\upbeta$). The difference vector $\vec{d}$ is decomposed in a state-independent translatory component $\vec{\kappa}$ and a heading-dependent component $\vec{\phi}$.}
    \label{fig:car_example}
    \vspace{-0.1cm}
\end{figure}

Each error component~$\vec{\phi}^i$ as well as~$\vec{\kappa}$ describe an individual error and conveys inherent characteristics, expressed by error parameters
The error parameters are summarized in the error state vector~$\vec{x}$, according to a defined error model (see also section \ref{sec:error_models}).
A peculiarity of each~$\vec{\phi}^i$ consists of dependence on a kinematic state.
In the case of multiple~$\vec{\phi}^i$, a specific kinematic state (e.g., heading angle) should only contribute to one error component in order not to risk observability.
Fig.~\ref{fig:car_example} shows an example of a situation with two erroneous position estimates that are decomposed into a state-independent component (translation) and a component $\vec{\phi}$ that depends on the heading angle~$\gamma$.

\subsubsection{Problem Statement}

Based on the preceding definitions, the error decomposition can be performed by comparing model $\operatorname{D}$ and real-world measurement $\vec{d}$.
However, true decomposition cannot be computed from single measurements, since it may result in an infinite amount of solutions.
For this reason, we observe the agent's dynamics over time, i.e., multiple time steps $j$.
More specifically, we determine the parameter set $\vec{x}^*$ that best fits the model according to
\begin{equation}
    \label{eq:diff_vec_argmin}
    \vec{x}^* = \arg \min_{\vec{x}} \frac{1}{m} \sum_{j=0}^m (\vec{d}_j - \operatorname{D}(\vec{x}, \vec{u}_j))^2 .
\end{equation}
A solution can be found by mean squared error minimization, according to section \ref{sec:filter}.

\section{The Error Model}
\label{sec:error_models}

\subsection{Error Model Description}
\label{sec:error_model_description}

In an error-free system, heterogeneous localizers provide similar position estimates.
However, in reality, position estimates do not align due to a variety of potential errors, ranging from sensor intrinsic problems or miscalibration to mapping-related issues.
In order to investigate and decompose such errors, a model is required. 
More precisely, in order to decompose a localization error according to equations (\ref{eq:diff_vec}) and (\ref{eq:diff_vec_argmin}), one model for each investigated error type, as well as a common reference frame needs to be defined.

Regarding the reference frame, especially the Cartesian navigation frame and the Cartesian body frame present suitable candidates since the calculation of distances becomes trivial in comparison to e.g., using polar coordinates.
While the navigation frame describes position with regard to a local tangent plane in east, north and up direction, the body frame is used to describe positions with regard to a fixed point on the vehicle in x (forward), y (left) and z (up) direction.
Without loss of generality, we choose the navigation frame as the main reference in subsequent explanations.

An error model, on the other hand, defines a mapping from a set of parameters and measured states into difference-vector space. 
To illustrate the error models, let us first consider two fundamental classes of errors that influence the localization error:
a) errors originating within the agent, and b) errors originating in the environment representation (e.g., due to mapping errors).

Errors originating from within the agent are not invariant under position changes of the mobile agent.
A common example of this is errors due to miscalibration of the localizer setup, i.e., the relative position of the localizers is not known or erroneous.
Such a translation can be described by the parameter vector $\vec{x}^\text{Agent}$.
In order to formulate the error, the body-fixed translation needs to be transformed to match the chosen reference frame, i.e., the navigation frame, under consideration of the orientation of the vehicle.
For the two-dimensional case, the error is modeled by,
\begin{equation}
    \vec{\phi}^\text{Agent}(\vec{x}^\text{Agent}, \gamma) = \begin{pmatrix}
            \cos(\gamma)	& -\sin(\gamma) \\
            \sin(\gamma) 	& \cos(\gamma)
    \end{pmatrix} \cdot \vec{x}^\text{Agent},
    \label{eq:phi_agent}
\end{equation}
with $\gamma$ being the heading angle of the agent.
This model can be extended for the three-dimensional case under additional consideration of the agent's roll and pitch angles and the corresponding rotation matrix.

\begin{figure}[ht!]
    \centering
	\includegraphics[width=0.98\linewidth]{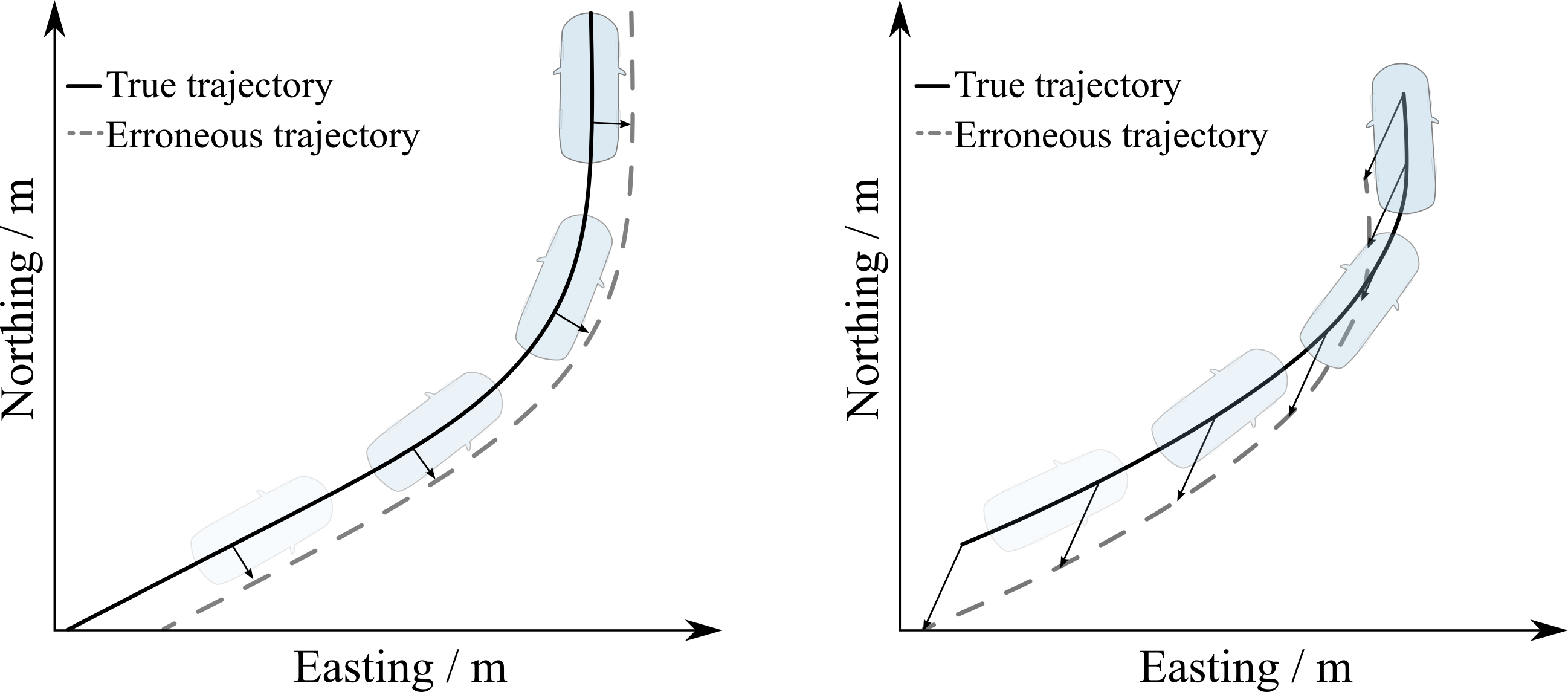}
    \caption{Comparison of a vehicle-inherent error, e.g., calibration (left), and an environment-inherent error, e.g., due to a map offset (right).}
    \label{fig:comparison}
    \vspace{-0.3cm}
\end{figure}

Next, we will discuss a second fundamental class of errors that originate in the environment's representation. 
The simplest example is a shifted map, expressed by a translation in the navigation frame.
In this case, the error model can be expressed by a parameter vector $\vec{x}^\text{Trans}$ which describes a 2-dimensional translation.
This presents the special case of an error model that is not dependent on any agent state and therefore, in accordance with (\ref{eq:diff_vec}), referred to as $\vec{\kappa}$.
The error is then modeled by 
\begin{equation}
    \vec{\kappa}(\vec{x}^\text{Trans}) = \vec{x}^\text{Trans}.
    \label{eq:kappa}
\end{equation}
In order to illustrate the two described error classes, Fig.~\ref{fig:comparison} visualizes the true trajectory as well as the erroneous trajectory for both cases.

A special case of environment-inherent errors presents position-dependent ones.
In this case, both positions, $\vec{p}_\upalpha$ and $\vec{p}_\upbeta$, need to be considered. 
A general recipe for developing a corresponding error model is as follows:

The first step is to select one of the two localizers as a reference, e.g., $\upalpha$ (can be any of the two).
Next, the position estimates $\vec{p}_\upbeta$ of the other localizer $\upbeta$ is expressed as a function of $\vec{p}_\upalpha$.
This is done under the assumption of transformation $\vec{e}$, modeling the expected relationship between $\vec{p}_\upalpha$ and $\vec{p}_\upbeta$ according to
\begin{equation}
	\vec{p}_\upalpha = \vec{e}( \vec{p}_\upbeta; \vec{x}^i) , 
\end{equation}
\begin{equation}
	\vec{p}_\upbeta = \vec{e}^{-1}( \vec{p}_\upalpha; \vec{x}^i).
\end{equation}
Here, $\vec{x}^i$ is a parameter vector that describes the transformation. 
Common examples of such deformations are a rotation, shearing, or scaling of the map, expressed by the corresponding transformation $\vec{e}$.

Assuming that there is no other error, i.e., all other components are zero, we can refer to equation (\ref{eq:diff_vector_measured}) by setting
\begin{equation}
	\vec{d} = \vec{\phi}^i(\vec{x}^i, \vec{p}_\upalpha) = \vec{p}_\upalpha - \vec{p}_\upbeta .
\end{equation}
Substituting $\vec{p}_\upbeta$ then yields 
\begin{equation}
	\vec{\phi}^i(\vec{x}^i, \vec{p}_\upalpha) = \vec{p}_\upalpha - \vec{e}^{-1}(\vec{p}_\upalpha; \vec{x}^i) , 
\end{equation}
which constitutes our model for a single error component. 
Here $\vec{x}^i$ is the parameter vector that corresponds to the error model $\vec{\phi}^i$.

We highlight that the preceding error models serve as examples and can be adapted or extended depending on the use case.
The illustrated error models address the usage of low-accuracy maps by identifying geometric transformations and target, amongst others, plug-and-play systems, which benefit from the calibration aspect.
More error models, considering similar or other inputs (e.g., time) and phenomena can be formulated according to the same scheme.

\subsection{Error Model Combination and Decomposability}
\label{sec:combination}

The models described above allow for a description of potential errors of a hybrid localization system.
By combining two or more  error models, we can model the difference vector according to (\ref{eq:diff_vec}).
Before designing an estimator, one has to consider the question if systems states can be recovered based  on the observation of the output.
In this context, observability describes whether latent variables can be identified based on the analysis of measurements and system outputs.
In general, a system is considered to be observable if all initial states $\vec{x}_0$ can be determined by analyzing the evolution of the input $\vec{u}$ and output $\vec{y}$ of a dynamic system within a limited time interval for all inputs.

A formal criterion for observability is given in \cite{JAdamy2018}.
Consider a system that is described by a vector differential equation defining the process according to 
\begin{equation}
    \begin{split}
        \vec{\dot{x}} &= \vec{f}(\vec{x},\vec{u}) \\
        y &= {g}(\vec{x},\vec{u}), 
    \end{split}
    \label{eq:sys_eq_miso}
\end{equation}
with $\vec{x} \in D_{\vec{x}} \subseteq \mathbb{R}^n$ being the state vector, $\vec{u} \in C_{\vec{x}} \subseteq C^{n-1}$ being the input vector,  $y$ being the output, and $\vec{f}$ being the process equation.

In the context of a nonlinear system according to (\ref{eq:sys_eq_miso}), Lie derivatives are used to derive an observability criterion.
In general, Lie derivatives are defined as the gradient of a scalar function multiplied by a vector field.
The repeated Lie derivative using the chain rule is defined as 
\begin{equation}
    \begin{split}
        {y} &= {g}(\vec{x}, \vec{u}) \\
        \dot{y} &= \mathcal{L}_{\vec{f}} {g} = \frac{\delta {g}}{\delta \vec{x}}\vec{f} + \frac{\delta {g}}{\delta \vec{u}}\dot{\vec{u}} = {h}_1(\vec{x}, \vec{u}, \dot{\vec{u}}) \\
        \ddot{y} &= \mathcal{L}_{\vec{f}} {h}_1 = \frac{\delta {h}_1}{\delta \vec{x}}\vec{f} + \frac{\delta {h}_1}{\delta \vec{u}}\dot{\vec{u}} + \frac{\delta {h}_1}{\delta \dot{\vec{u}}}\ddot{\vec{u}} = {h}_2(\vec{x}, \vec{u}, \dot{\vec{u}}, \ddot{\vec{u}}) \\
        &\vdots   \\
        y^{(n-1)} &= \mathcal{L}_{\vec{f}} {h}_{n-2} = {h}_{n-1}(\vec{x}, \vec{u}, \dot{\vec{u}}, \ddot{\vec{u}}, \dots, {\vec{u}^{(n-1)}}) ,
    \end{split}
    \label{eq:lie_set}
\end{equation}
which can be rewritten as
\begin{equation}
    \begin{aligned}
        \vec{z} = \begin{pmatrix}
        y \\
        \dot{y} \\
        \ddot{y} \\
        \vdots \\
        \overset{(n-1)}{y}
        \end{pmatrix} &= \begin{pmatrix}
        \operatorname{g}(\vec{x}, \vec{u}) \\
        \operatorname{h}_1(\vec{x}, \vec{u}, \dot{\vec{u}}) \\
        \operatorname{h}_2(\vec{x}, \vec{u}, \dot{\vec{u}}, \ddot{\vec{u}}) \\
        \vdots \\
        \operatorname{h}_{n-1}(\vec{x}, \vec{u}, \dot{\vec{u}}, \ddot{\vec{u}}, \dots, {\vec{u}^{(n-1)}})
        \end{pmatrix} \\
        &=  \operatorname{q}(\vec{x}, \vec{u}, \dot{\vec{u}}, \ddot{\vec{u}}, \dots,{\vec{u}^{(n-1)}}).
    \end{aligned}
    \label{eq:lie_rewritten}
\end{equation}
According to \cite{JAdamy2018}, the described nonlinear system (\ref{eq:sys_eq_miso}) is observable if solving equation (\ref{eq:lie_rewritten}) leads to a unique solution for all $\vec{x} \in D_{\vec{x}} \subseteq \mathbb{R}^n$ and $\vec{u} \in C_{\vec{x}} \subseteq C^{n-1}$.

From this criterion, we now draw conclusions for the decomposability of our error-motivated system.
In terms of a state-space representation, the system can be described by
\begin{equation}
    \label{eq:sys_eq_vd}
    \begin{split}
        \vec{\dot{x}} &= \vec{f}(\vec{x},\vec{u}) = 0 \\
        \vec{y} &= \vec{g}(\vec{x},\vec{u}) = \operatorname{D}(\vec{x}, \vec{u}) = \vec{\kappa}(\vec{x}) + \sum_i \vec{\phi}^i(\vec{x}, \vec{u}),
    \end{split}
\end{equation}
with the measurable output vector $\vec{y}$ being the difference model from equation (\ref{eq:diff_vec}).
At this point, we highlight that input $\vec{u}$ represents the kinematic state of an agent that shall not be confused with the state $\vec{x}$ of the system, i.e., the parameters of the error model.  

Comparing equations~\eqref{eq:sys_eq_vd} and~\eqref{eq:sys_eq_miso}, two aspects need to be noted. 
First, the states of the system of equation (\ref{eq:sys_eq_vd})  are the static parameters of $i$ individually chosen error models. 
Therefore, state equation $\vec{f}(\vec{x},\vec{u})$ results in a zero vector which allows for a simplification of (\ref{eq:lie_set}), i.e., all summands containing  $ \vec{f}(\vec{x},\vec{u})$ vanish.

Second, the described criterion for system observability, according to \cite{JAdamy2018}, targets systems with multiple-input-single-output (MISO) behavior.
In the case of our decomposition system, we have a multiple-input-multiple-output (MIMO) system, where the output matches the dimensionality of the difference vector $\vec{d}$.
However, the requirement of being able to uniquely solve for the state still applies.

Following the initial argumentation that observability requires clear distinguishability of various initial states based on the evaluation of inputs and output measurements, we derive that one can successfully decompose a difference vector by estimating its parameters if the system of error components is solvable for the state.

We illustrate the theoretical considerations by using the example of an agent that is equipped with a miscalibrated hybrid system (expressed by a vehicle-inherent translation). 
Furthermore, the agent localizes itself based on a uniformly shifted two-dimensional map.
Therefore, we combine the corresponding equations (\ref{eq:phi_agent}) and (\ref{eq:kappa}) into a single model. We obtain

\begin{equation}
    \operatorname{D} = \vec{\phi}^\text{Agent}(\vec{x}^\text{Agent},\gamma) + \vec{\kappa}(\vec{x}^\text{Trans}).
    \label{eq:phikappa}
\end{equation}

Splitting equation (\ref{eq:phikappa}) according to equations  (\ref{eq:phi_agent}), (\ref{eq:kappa}) and computing the first Lie derivative for each leads to

\begin{equation}
\begin{split}
    d_{1} &= \vec{x}^\text{Agent}_{1} \cos{\left(\gamma \right)} - \vec{x}^\text{Agent}_{2} \sin{\left(\gamma \right)} + \vec{x}^\text{Trans}_{1} \\
    d_{2} &= \vec{x}^\text{Agent}_{1} \sin{\left(\gamma \right)} + \vec{x}^\text{Agent}_{2} \cos{\left(\gamma \right)} + \vec{x}^\text{Trans}_{2} \\
    \dot{d}_{1} &= \dot{\gamma} \left(- \vec{x}^\text{Agent}_{1} \sin{\left(\gamma \right)} - \vec{x}^\text{Agent}_{2} \cos{\left(\gamma \right)}\right) \\
    \dot{d}_{2} &= \dot{\gamma} \left(\vec{x}^\text{Agent}_{1} \cos{\left(\gamma \right)} - \vec{x}^\text{Agent}_{2} \sin{\left(\gamma \right)}\right).
\end{split}
\end{equation}

To further discuss the observability, we need to solve this system of four equations for the parameters $\vec{x}^\text{Agent}$ and $\vec{x}^\text{Trans}$.
By doing so we get:
\begin{equation}
\begin{split}
    \vec{x}^\text{Agent}_1 &= \frac{-\dot{d}_1 \sin(\gamma) + \dot{d}_2 \cos(\gamma)}{\dot{\gamma}}\\
    \vec{x}^\text{Agent}_2 &= -\frac{\dot{d}_1 \cos(\gamma) + \dot{d}_2 \sin(\gamma)}{\dot{\gamma}} \\
    \vec{x}^\text{Trans}_1 &= d_1 - \frac{\dot{d}_2}{\dot{\gamma}} \\
    \vec{x}^\text{Trans}_2 &= d_2 - \frac{\dot{d}_1}{\dot{\gamma}}.
\end{split}
\label{eq:phikappa_sol}
\end{equation}

We can see that both error model parameters depend on the input $\gamma$.
From the influence of the heading angle's turn rate on equations (\ref{eq:phikappa_sol}), we derive that the angle is not allowed to be constant for the parameters to be identified.
This means that the agent has to be moving, or rather turning, to decompose the error, which follows the expectations so far.

In a nutshell, the criteria presented in this section serve to determine beforehand if the combination of two or more error components promises successful decomposition results.

\subsection{Filter Selection}
\label{sec:filter}

The last step addresses the selection of a suitable estimator for identifying the latent variables, i.e., error parameters that cannot be directly measured.
In order to decompose the error vector, we are looking for an estimator that works online and allows us to embed our model to bias the estimation process towards our assumptions about the error's behavior.
While highlighting that our concept is not bound to a specific estimator, we propose to use a recursive Bayesian estimation method (Bayes Filter).
More specifically, we choose to employ a solution from the family of Kalman filters which is used  to identify  and optimize the parameter vector $\vec{x}$ by minimizing the least squares error according to equation (\ref{eq:diff_vec_argmin}).
Kalman Filtering follows a  prediction-update scheme that considers zero-mean Gaussian noises for the process as well as observations, expressed by covariance matrices $\mat{Q}$ and $\mat{R}$ respectively.
Considering our system as described in (\ref{eq:sys_eq_vd}), we look at a non-linear system.
While the Kalman Filter has been invented to tackle linear systems  \cite{RKalman1960}, extensions have been introduced to apply the concept to non-linear systems.
The most crucial difference is the description of measurements and processes by vector functions instead of matrices.
We suggest the usage of the Unscented Kalman Filter (UKF) as an online parameter estimator that handles non-linearities while not requiring us to compute and linearize the derivatives of the system by introducing a sampling technique that conserves the Gaussian character.
In brief, the Unscented Kalman filter tries to reconstruct the Gaussian character despite the non-linear behavior by using so-called sigma points, as proposed in \cite{WanMerwe2000}.
The non-linear vector function $\vec{f}$ is applied to a set of weighted sigma points, including the prior mean, from which a Gaussian distribution is then reconstructed in order to be compliant with the ideas of the original Kalman Filter. 

With regard to our system, the measurements for the filter constitute a sequence of difference vectors according to equation (\ref{eq:diff_vector_measured}) while the input is the required information on the vehicle dynamics such as the vehicle heading $\gamma$. 
The state space, estimated by the UKF, consists of the error parameters $\vec{x}^i$, stacked to one vector $\vec{x}$. 

Last but not least, let us have a closer look at the measurements.
The filter benefits from the estimated error variances of the preconnected localization components.
In the case of a Kalman filter, the inputs to the filter are assumed to be Gaussian random processes.
We consider the outputs of the two localization components $\vec{p}_\upalpha$ and $\vec{p}_\upbeta$ to be normally distributed random variables with $\vec{\Sigma}$ being the individual variance.
Therefore, we derive the difference vector
\begin{equation}
    \begin{split}
    	\vec{d} &\sim \vec{p}_{\upalpha} - \vec{p}_{\upbeta} \\
    			 &\sim \mathcal{N}(\vec{p}_{\upalpha}, \vec{\Sigma}_{\upalpha}) - \mathcal{N}(\vec{p}_{\upbeta}, \vec{\Sigma}_{\upbeta})   \\
    			 &\sim \mathcal{N}(\vec{p}_{\upalpha}, \vec{\Sigma}_{\upalpha}) + \mathcal{N}(-\vec{p}_{\upbeta}, \vec{\Sigma}_{\upbeta}) \\
    			 &\sim \mathcal{N}(\vec{p}_{\upalpha} - \vec{p}_{\upbeta}, \vec{\Sigma}_{\upalpha} + \vec{\Sigma}_{\upbeta}).
    \end{split}
    \label{eq:meas_co}
\end{equation}
Based on this, the measurement covariance matrix is set to $\mat{R}=\vec{\Sigma}_{\upalpha} + \vec{\Sigma}_{\upbeta}$, which is adapted at runtime according to the sensor processing. 

\section{Applicability and Evaluation}
\label{sec:error_application}

\subsection{Application Examples}
\label{sec:application}

Previous sections introduced error models and decomposability criteria on a theoretical level.
Of course, choosing the correct error model requires having a basic assumption of the characteristics of dominant errors. Therefore, the following section briefly outlines potential application areas by transforming abstract definitions into real-world examples.

\subsubsection{Predictive Localization}

In general, decomposing an error into its components based on an underlying model allows for anticipating the evolution of the error based on measured states (e.g., heading angle).
For example, separating an observable error into an orientation-dependent and an orientation-independent part can be beneficial in estimating an agent's position in non-straight-driving situations.
Only an error vector, e.g., currently facing east, is detectable without decomposition.
After successful decomposition, predicting how the error evolves when the agent's heading angle is altered, e.g., turning at an intersection, becomes possible.

\subsubsection{Map Correction}

Several exemplary error models have been presented in section \ref{sec:error_models}, from which two or more can be chosen and combined based on the underlying real-world problem.
Considering environment-inherent error models, the identified errors are commonly induced by localization modules that work in a global reference frame, e.g., GNSS, or by the map, e.g., a translation or rotation.
While a broad range of research deals with the latter case, we tackle the first one or a combination of both. 
Assuming the errors induced by the absolute localization modules are corrected or have a mean of zero over time or over repeated measurements, identified, environment-dependent error parameters can be used to improve the map. 
Note that translations of a map can be regional; for example, it is a common phenomenon that individual road segments of publicly available maps, such as those derived from OpenStreetMap, are shifted due to inconsistent placement of the road center.
The decomposition allows for detecting offsets of such segments. 
However, in the case of segment-wise offsets, it is suggested that a new filtering process be started when leaving a segment while previously acquired information on other error sources can be incorporated as an initial estimate.

\subsubsection{Hybrid Localization Calibration}

While the calibration of homogeneous networks of static sensors has been researched before (see, e.g., \cite{AshMoses2008}), the Kappa-Phi method also allows for calibrating heterogeneous sensor setups mounted on a moving agent.
In this context, we consider vehicle-inherent errors.
According to section \ref{sec:hybrid_loc_sys}, each localization component is defined by a logical center, while the relative distance between both logical centers and the vehicle's reference point, e.g., the center of the rear axis, should be known.
If this is not the case, calibration of the hybrid localization sensor setups is needed.
More specifically, calibration describes the geometric transformation between two modules.
Again, we stress that we focus on the calibration with regard to the relation between localizers $\upalpha$ and localizer~$\upbeta$.
Reversing the conclusion means that the error decomposition allows us to calibrate one module if we know the position (and orientation) of the other one.

One tangible example is a plug-and-play system. 
In this case, the computing hardware (e.g., handheld, vehicle computer, mobile phone) is connected to all sensors. 
In contrast, the GNSS sensor is installed in the vehicle, and the camera is mounted randomly on demand (fixed to form one rigid body). 
Changing the position of one component is also possible since it could be considered in the error estimation process by resetting the estimate of the orientation-dependent error.
However, if one wants to draw conclusions with regard to the center of the vehicle, then the position of at least one of the two modules has to be known. 

\subsection{Simulations}
\label{sec:experiments}

Last but not least, the theoretic considerations are demonstrated by exemplary experiments.
We want to stress that the evaluations primarily illustrate the concept and show potential for application. 

\begin{figure}[ht!]
    \centering
	\includegraphics[width=1.0\linewidth]{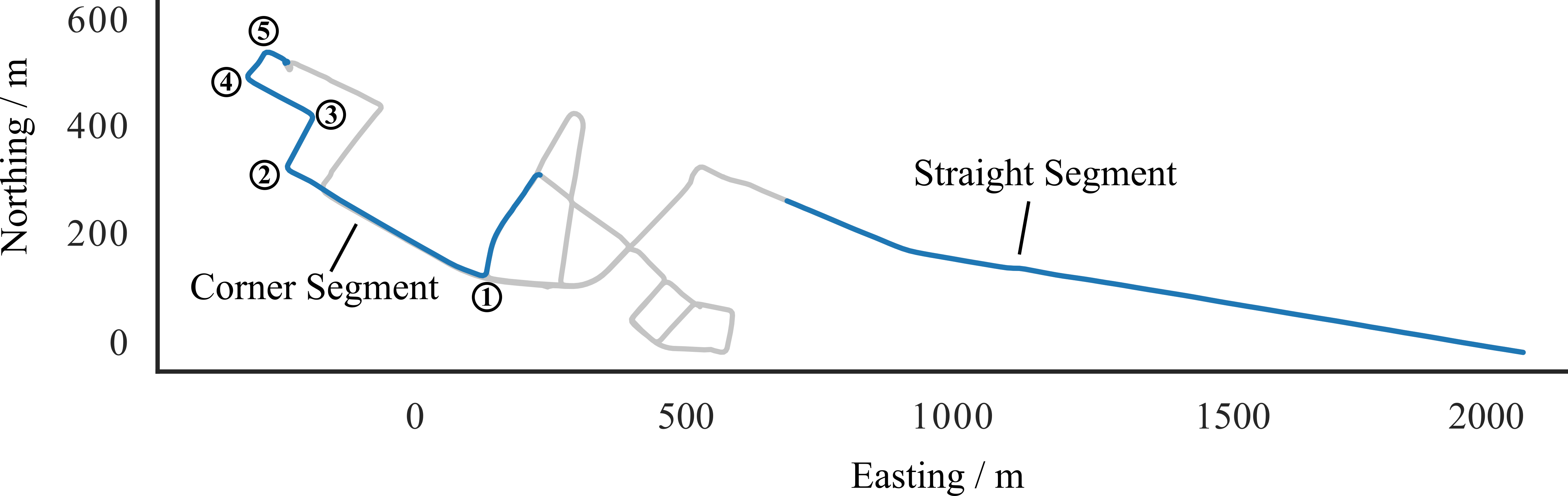}
    \caption{GNSS path with two segments highlighted: a mostly straight segment as well as a segment with five turning situations.}
    \label{fig:gnss}
    \vspace{-0.2cm}
\end{figure}

Consider again the example of an agent with a miscalibrated hybrid sensor setup that uses a map with an offset, as described in section \ref{sec:combination}.
A GNSS trace is recorded for the simulation, depicted in Fig. \ref{fig:gnss}.
To simulate a vehicle-inherent miscalibration, a vehicle-fixed translation of $\vec{e}^\text{Agent}=(2~m, 1~m)$ in the forward and left direction, with additional normally distributed noise (standard deviation of 0.2~m) is added  in every simulation step.
Furthermore, an orientation-independent offset $\vec{e}^\text{Trans}=(3~m, 2~m)$ in the east and north directions is added to simulate the map offset.

\begin{figure*}[ht!]
    \centering
    \vspace{0.2cm}
	\includegraphics[width=1\linewidth]{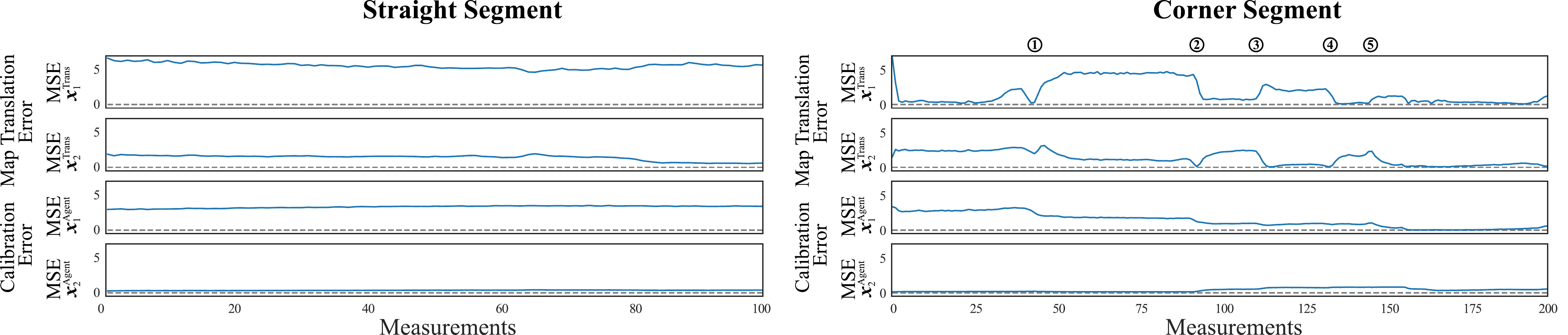}
    \caption{Evaluation of the mean squared error (MSE) for a straight segment (left) and for a segment with several turning points (right). Since the decomposition of the present error model depends on alternating heading angles, the errors of the four parameters converge for the corner segment only.
    The encircled numbers refer to positions on the path of Fig.~\ref{fig:gnss}.}
    \label{fig:eval_2}
    \vspace{-0.35cm}
\end{figure*}

According to section \ref{sec:combination}, the given combination of error models expects an alteration of the input, i.e., the agent's heading angle, to be able to decompose the error successfully.
To investigate this further, two segments of the GNSS trace are selected:
a) a mostly straight segment of 100 sample points (straight segment), and b) a segment with five turning situations at intersections of 200 sample points (corner segment).  
For the simulation, an Unscented Kalman Filter is used with an initial state estimate of $\vec{x}=(0,0,0,0)$, initial uncertainty in filtered states $\vec{P}=\operatorname{diag}(10)$ and all entries of the process noise $\vec{Q}$ set to 0.1.
The covariance matrix $\vec{R}$ of the measurement noise is set according to (\ref{eq:meas_co}).

Fig. \ref{fig:eval_2} shows the simulation results.
For each case, 100 runs are performed.
In the case of a straight segment (left), none of the four parameters ($\vec{x}^\text{Agent}_{1},\vec{x}^\text{Agent}_{2},\vec{x}^\text{Trans}_{1},\vec{x}^\text{Trans}_{2}$) is converging, expressed by the mean squared error (MSE).
On the other hand, the error is converging in the case of the corner segment. 
Significant jumps in the error can be seen at every turning point (indicated by an encircled number).
In contrast, the error converges towards the desired state of all four MSEs being zero.

Since statistical methods are used for the error estimation, we obtain a certainty measure and the error estimate. 
Both estimates and their certainty can be shared since every vehicle's map errors are the same. 
Furthermore, repeated trips by the same or multiple agents  improve the estimate of the error parameters.

\section{Conclusions}
\label{sec:conclusion}
 
In this paper, we presented a general approach to decomposing errors within a hybrid localization system.  
More specifically, the novel Kappa-Phi method enables simultaneous plausibilization of disparities between viewpoint and absolute localization processes into a sum of parameterized functions of the measured state.
This is done by leveraging the individual characteristics of the underlying heterogeneous localization methods.
We believe that the resulting information from exploiting these disparities between localization results can be beneficial for tasks such as lane-level navigation by the prediction of future disparities, correction of systematic map errors, and calibration of the hybrid system.
In this context, an Unscented Kalman Filter is used and preliminary evaluations have been run on recorded data with artificial errors.
One highlighted use case presents the example of combining absolute and viewpoint localization without the restraint of requiring highly accurate map data.

\section{Future Research}

The next step is to evaluate the performance of the method in the context of real-world errors. 
Furthermore, the applicability of the approach towards additional error models, such as time-dependent ones, e.g., concerning IMU drift, should be investigated. 
An upcoming task is the implementation of the approach on a test vehicle in order to evaluate the influence of further effects, such as real-world noise.
In that regard, more complex errors within the map, e.g., of topological nature or occlusions, need to be considered as well. 
It is also subject to subsequent evaluations of how the system performs with non-constant, slowly changing, and jumping error parameters.
Overall, the first evaluations illustrated the concept, and we believe that the proposed approach presents promising components for future localization architectures.

\section*{Acknowledgment}

The authors would like to thank Axel Koppert from OHB Digital Solutions GmbH, Graz, as well as Tim Schoonbeek from the Eindhoven University of Technology, the Netherlands, for their support.  
Map data \copyright~OpenStreetMap contributors, licensed under the Open Database License (ODbL) and available from http://www.openstreetmap.org.

\balance

\bibliographystyle{IEEEtran}
\bibliography{main}

\end{document}